\documentclass[sigconf]{acmart}

\usepackage{multirow}

\usepackage{lipsum}

\AtBeginDocument{%
	}

\copyrightyear{2023}
\acmYear{2023}
\setcopyright{acmlicensed}
\acmConference[MM '23]{Proceedings of the 31st ACM International Conference on Multimedia}{October 29--November 3, 2023}{Ottawa, ON, Canada.}
\acmBooktitle{Proceedings of the 31st ACM International Conference on Multimedia (MM '23), October 29--November 3, 2023, Ottawa, ON, Canada}
\acmPrice{15.00}
\acmISBN{979-8-4007-0108-5/23/10}
\acmDOI{10.1145/3581783.3612128}

\newcommand\blfootnote[1]{%
\begingroup
\renewcommand\thefootnote{}\footnote{#1}%
\addtocounter{footnote}{-1}%
\endgroup
}

\begin{document}
        \title{Unfolding Once is Enough: A Deployment-Friendly Transformer Unit for Super-Resolution}

        \author{Yong Liu$^{1,2,*}$, Hang Dong$^{3,*,\dagger}$, Boyang Liang$^{4,*,\ddagger}$, Songwei Liu$^{3}$, Qingji Dong$^{1,2,\ddagger}$, \\Kai Chen$^{3}$, Fangmin Chen$^{3}$, Lean Fu$^{3}$, Fei Wang$^{1,2}$}
        \affiliation{
         \institution{
            $^1$ National Key Laboratory of Human-Machine Hybrid Augmented Intelligence\\
            $^2$ IAIR, Xi’an Jiaotong University\\
            $^3$ ByteDance Inc\\
            $^4$ School of Software Engineering, Xi’an Jiaotong University
        }
        \city{}
        \country{}
        }
        \email{liuy1996v@qq.com, dhunter1230@gmail.com, 951240992@qq.com}
		\renewcommand{\shortauthors}{Yong Liu et al.}

	\begin{abstract} 
		Recent years have witnessed a few attempts of vision transformers for single image super-resolution (SISR). 
        Since the high resolution of intermediate features in SISR models increases memory and computational requirements, efficient SISR transformers are more favored. 
        Based on some popular transformer backbone, many methods have explored reasonable schemes to reduce the computational complexity of the self-attention module while achieving impressive performance. 
        However, these methods only focus on the performance on the training platform (e.g., Pytorch/Tensorflow) without further optimization for the deployment platform (e.g., TensorRT). 
        Therefore, they inevitably contain some redundant operators, posing challenges for subsequent deployment in real-world applications. 
		In this paper, we propose a deployment-friendly transformer unit, namely UFONE (i.e., UnFolding ONce is Enough), to alleviate these problems. 
		In each UFONE, we introduce an Inner-patch Transformer Layer (ITL) to efficiently reconstruct the local structural information from patches and a Spatial-Aware Layer (SAL) to exploit the long-range dependencies between patches. 
		Based on UFONE, we propose a Deployment-friendly Inner-patch Transformer Network (DITN) for the SISR task, which can achieve favorable performance with low latency and memory usage on both training and deployment platforms. 
        Furthermore, to further boost the deployment efficiency of the proposed DITN on TensorRT, we also provide an efficient substitution for layer normalization and propose a fusion
        optimization strategy for specific operators. 
		Extensive experiments show that our models can achieve competitive results in terms of qualitative and quantitative performance with high deployment efficiency. 
        Code is available at \url{https://github.com/yongliuy/DITN}.
	\end{abstract}



	\maketitle
 
        \blfootnote{*Contributed equally.}
        \blfootnote{$\dagger$Corresponding author.}
        \blfootnote{$\ddagger$Work done during an internship at ByteDance Inc.}
        \vspace{-1cm}

 	\begin{figure}[h]
        \setlength{\abovecaptionskip}{0pt}
        \setlength{\belowcaptionskip}{-1pt}
		\centering
		\includegraphics[width=0.9\linewidth]{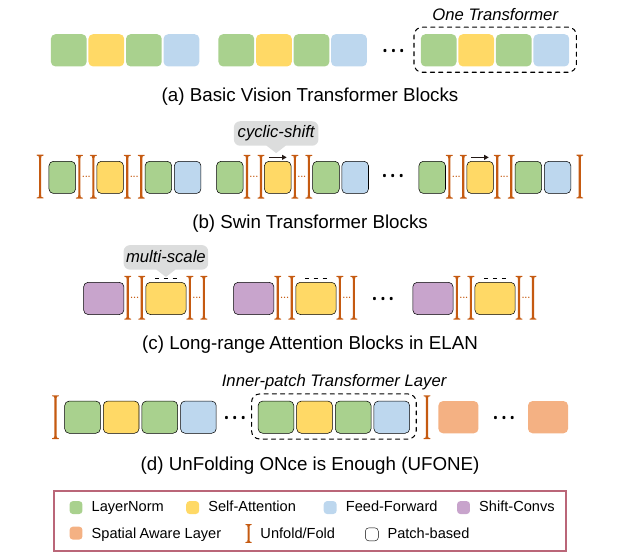}
		\caption{The structure of various vision transformers. (a) The basic vision transformer block performs global self-attention, which is unacceptable for large image sizes; (b) The Swin transformer block \cite{liu2021swin,liang2021swinir} adopts window-based self-attention, but requires extensive reshaping operations to fold-and-unfold features for cyclic-shifts; (c) The long-range attention block in ELAN \cite{zhang2022efficient} divides features into multiple patch sizes and computers self-attention separately, which still involves a large number of reshaping operations; (d) The proposed UFONE unfolds/folds features only once and transfers all the transformer components to the patch level, which greatly reduces the computational cost and minimizes the reshaping operation. }
		\label{transformers}
	\end{figure}
 
	\section{Introduction}

	Single Image Super-Resolution (SISR) \cite{glasner2009super} aims to reconstruct a high-resolution (HR) image from its low-resolution (LR) observation, 
	which has received increasing research attention in multimedia and computer vision communities for decades. 
        With the rapid development of convolutional neural networks (CNN), various CNN-based SISR models \cite{zhang2018residual,hui2019lightweight,li2020lapar,luo2020latticenet,xiao2021self,li2021information,zhang2021pffn} have been proposed and have achieved consistent performance improvements. 
        However, convolution filters struggle with extracting adaptive features for the dynamic inputs due to their fixed weights at inference \cite{liang2021swinir,zamir2022restormer}, which limits the performance of CNN-based methods.

	To address this problem, a series of vision transformers have been introduced for low-level tasks \cite{yang2020learning,chen2021pre,tsai2022stripformer,song2023vision}, including SISR. 
	However, the computational complexity of self-attention in transformers grows quadratically with the feature map size, resulting in an unacceptable computational cost for SISR tasks. 
	From an efficient perspective, Liu \textit{et al.} \cite{liu2021swin} proposed a window-based Swin Transformer, which computes self-attention on non-overlapping windows and build connections with nearby windows through a cyclic-shift mechanism. 
	Based on this, Liang \textit{et al.} \cite{liang2021swinir} proposed a baseline image restoration model, namely SwinIR, which uses a stack of residual Swin Transformer blocks to extract deep features. 
        To compensate for the weak long-range dependencies that may be generated by the cyclic-shift mechanism, Zhang \textit{et al.} \cite{zhang2022efficient} further developed an efficient long-range attention network (ELAN), which applies a group-wise multi-scale self-attention to efficiently capture long-range dependencies. 
        Although these methods have significantly reduced the computational cost, they need to alternately unfold features into flat patches for multi-scale fusion and fold them for cyclic-shift, which inevitably introduces a large number of reshape operators, as shown in Figures \ref{transformers} (b) and (c). 
        Since the reshape operator will not be counted in the FLOPs and parameter computations, most existing methods ignore its adverse impact on the deployment of transformer models. 
        In fact, reshape operators can be treated as "glue operators" during the deployment and extensive reshaping operations will bring computational burden and cause more memory usage during the inference stage, especially for the deployment models for dynamic inputs. 
        Therefore, there is still room for further improvement in light-weighted SISR transformer and its deployment efficiency.

\begin{figure}[t]
        \setlength{\abovecaptionskip}{0pt}
        \setlength{\belowcaptionskip}{-1pt}
		\centering
		\includegraphics[width=\linewidth]{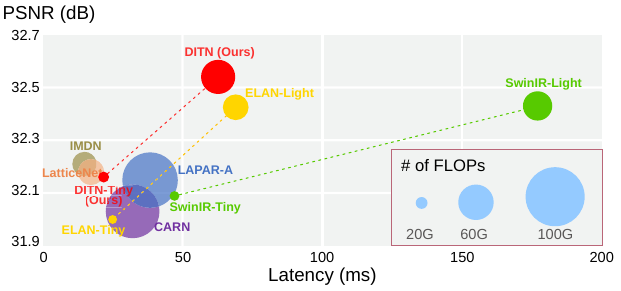}
		\caption{Performance comparison of existing light-weighted SISR methods (< 2M parameters) on Set5 \cite{bevilacqua2012low} ($\times$4). The Swin-Tiny, ELAN-Tiny, and DITN-Tiny are obtained by reducing the corresponding model parameters to 1/3 times.}
		\label{visualization}
	\end{figure}

       Inspired by the observation that the reconstruction of local image details and textures plays a more important role for SISR than high-level tasks \cite{yoo2020rethinking,ning2021uncertainty}, we propose a Deployment-friendly Inner-patch Transformer Network (DITN) to reduce the amount of reshape operators.
        As shown in Figure \ref{transformers}(d), the proposed DITN transfers all the transformer layers to the patch level and develops a deployment-friendly transformer unit UFONE (i.e., UnFolding ONce is Enough), in which feature maps are unfolded only once in the beginning and successive Inner-patch Transformer Layers (ITL) are performed before the folding operation.  
        In this way, we can not only leverage the content-adaptive flexibility of transformer to recover rich local structural information from patches but also can minimize the reshaping operations. 
        To capture potentially useful long-range dependencies that are neglected in ITL, we further propose a Spatial-Aware Layer (SAL) which combines the efficient MetaFormer-style structure \cite{yu2022metaformer} and dilated convolutions \cite{wang2018smoothed} with large receptive fields. 
	Both quantitative and qualitative results on the Pytorch framework show that our DITN achieves state-of-the-art performance compared to other light-weighted SISR methods, which can be seen visually in Figure \ref{visualization}. 
        In addition, to boost the efficiency of the proposed DITN on deployment platform (e.g., TensorRT), we further propose a deployment version of the DITN for realistic scenario by replacing the layer normalization (LayerNorm) and propose an effective operator fusion optimization strategy for specific operators of the self-attention module. 
	Experiments on both training and deployment platforms demonstrate that our model can achieve extremely low latency on both static and dynamic inputs, while still maintaining excellent results. 

	In summary, the main contributions of this paper are as follows: 
	\begin{itemize}
		\item We propose a Deployment-friendly Inner-patch Transformer Network for SISR, namely DITN, which can not only exploit the content-adaptive flexibility of transformer to recover rich local structural information from patches but also can minimize the reshaping operations.

		\item We further boost the deployment efficiency of DITN by replacing the LayerNorm and  exploring a fusion optimization strategy for specific operators on TensorRT. All the modifications can be easily transplanted to other transformer-based models.

		\item We verified the outstanding performance of DITN on various standard benchmark datasets and different platforms. To the best of our knowledge, this is the first work to optimize SISR transformer from the view of deployment scenarios and achieve high efficiency on the training platform.

	\end{itemize}
	

		\begin{figure*}[t]
        \setlength{\abovecaptionskip}{0pt}
        \setlength{\belowcaptionskip}{-1pt}
		\centering
		\includegraphics[width=\linewidth]{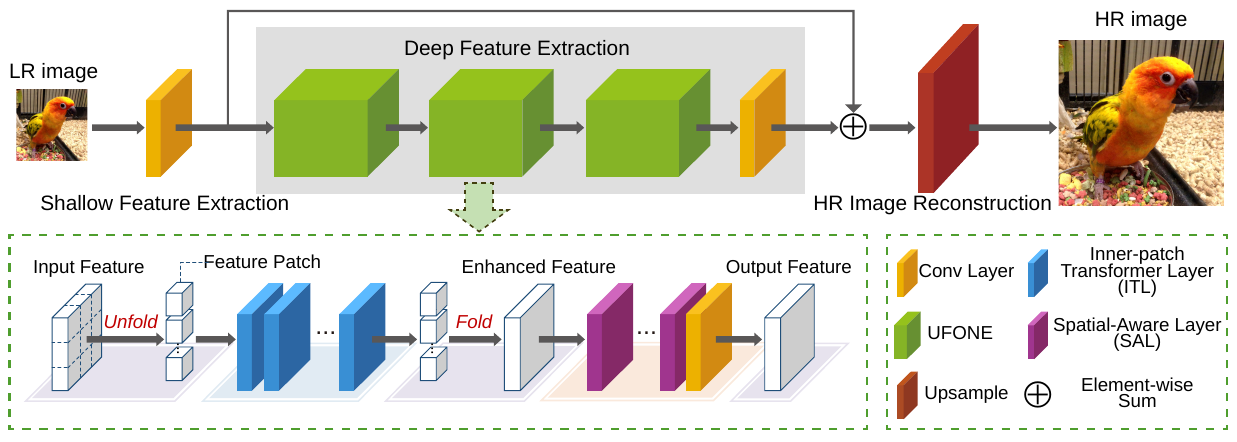}
		\caption{The overall network architecture of the proposed DITN.}
		\label{network}
        \vspace{-1em}
	\end{figure*}

	\section{Relate work}
	\subsection{Single Image Super-Resolution}
        Traditional methods mainly use CNN-based network architectures to find feasible SISR solutions \cite{dong2014learning,tai2017image,wang2018esrgan,zhang2018residual,chen2019joint,li2021information}, which includes a series of light-weighted methods \cite{ahn2018fast,hui2019lightweight,li2020lapar,muqeet2020ultra,luo2020latticenet,xiao2021self,zhang2021pffn}. 
        All these methods have made varying degrees of progress. 
         However, adaptive feature extraction for the dynamic inputs always fails due to the fixed weights of convolution filters at inference \cite{liang2021swinir,zamir2022restormer}. 
        Recently, some efficient transformer-based methods have been proposed to address this issue. 
	For example, 
	Liang \textit{et al.} \cite{liang2021swinir} introduced an image restoration model, SwinIR, based on the Swin Transformer \cite{liu2021swin}. 
	SwinIR utilizes a stack of Residual Swin Transformer Blocks to extract deep feature information, each RSTB consists of Swin Transformer layers, a convolutional layer, and a residual connection. 
 	Zhang \textit{et al.} \cite{zhang2022efficient} proposed an efficient long-range attention SR network (ELAN), which employs shift convolutions to extract the local image structural information and uses group-wise multi-scale self-attention modules to exploit the long-range image dependency. 
	Lu \textit{et al.} \cite{lu2022transformer} proposed an Efficient Super-Resolution Transformer (ESRT). 
	ESRT first uses a light-weighted CNN backbone to extract deep features, and then uses a light-weighted transformer backbone to capture long-range dependencies between similar feature regions. 
	While promising results have been achieved, these methods contain extensive reshape operators in the transformer models, which results in computational burden and increased memory usage during the inference stage, especially for dynamic inputs on the deployment platform. 
	\vspace{-2mm}
	
	\subsection{Deployment-oriented Transformer}
        To bridge the gap between the training platform and the deployment platform of the transformer-based models, 
        Li \textit{et al.} \cite{li2022next} proposed a next generation vision transformer (Next-ViT) to strike a balance between latency and accuracy on the deployment platform. 
        Xia \textit{et al.} \cite{xia2022trt} presented a family of TensorRT-oriented transformers, namely TRT-ViT, to compensate for the performance problem deployed on TensorRT. 
        In addition, Tang \textit{et al.} \cite{tang2023elasticvit} introduced ElasticViT, a deployment-friendly method for training a high-quality vision transformer supernet. 
        Although these methods have made varying degrees of progress in realistic industrial deployments, they have mostly developed for high-level tasks such as image classification, semantic segmentation, and object detection, which are quite different from SISR. 
        On the one hand, SISR is sensitive to latency due to the the high resolution of the intermediate features. 
        On the other hand, SISR focuses mainly on reconstructing local structural details of images, rather than obtaining key semantic information from the global feature as in high-level tasks \cite{yoo2020rethinking,ning2021uncertainty}.

        Therefore, the strategies mentioned above for high-level tasks are not applicable to SISR. 
        For this reason, we explore a deployment-friendly efficient network for SISR in this paper.

	
	\section{Proposed Method}

	\begin{figure*}[t]
        \setlength{\abovecaptionskip}{0pt}
        \setlength{\belowcaptionskip}{-1pt}
		\centering
		\includegraphics[width=\linewidth]{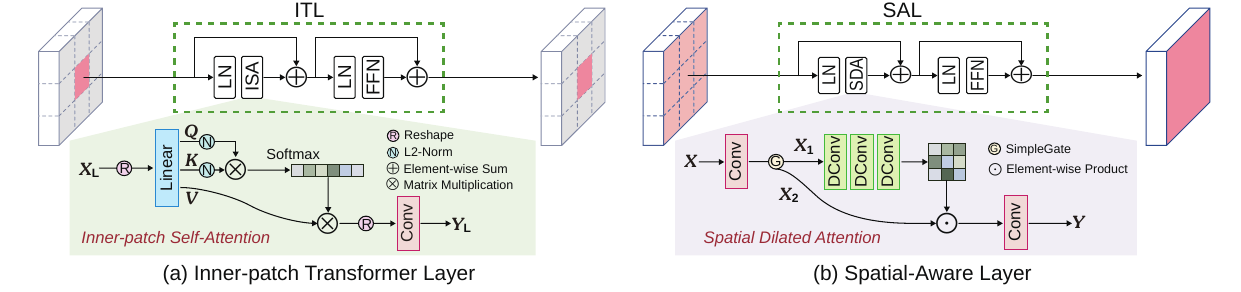}
		\caption{The details of the proposed Inner-patch Transformer Layer (ITL) and Spatial-Aware Layer (SAL).}
		\label{details}
        \vspace{-1em}
	\end{figure*}

	\subsection{Overall Architecture}
	The overall architecture of the proposed DITN is illustrated in Figure \ref{network}, which consists of three modules: shallow feature extraction, deep feature extraction and HR image reconstruction modules. 
	The input of DITN is an LR image $ I_{LR} \in \mathbb{R}^{3\times H\times W} $, where $ H $ and $ W $ are the height and width of the LR image, respectively. 
	We first pass $ I_{LR} $ into a shallow feature extraction module, which consists of only a convolution layer, to extract the shallow feature: 
	\begin{equation}
		F_{SF} = ConV(I_{LR}),  
	\end{equation} 
	where $ F_{SF} \in \mathbb{R}^{C\times H\times W} $ and $ C $ is the channel number of the intermediate feature. 
	The shallow feature is then fed into a deep feature extraction module with $ K $ deployment-friendly transformer units (UFONE) and a convolution layer. 
	That is: 
	\begin{equation}
		F_{DF} = ConV(UFONEs(F_{SF})), 
	\end{equation} 
	where $ F_{DF} $ represents the corresponding deep feature and $ F_{DF} \in \mathbb{R}^{C\times H\times W} $. 
	The convolution layer at the end of the deep feature extraction module can not only bring the inductive bias of convolution into UFONEs, but also facilitate further feature fusion of shallow and deep features through a residual connection. 
	Finally, the features are passed to the HR image reconstruction module (i.e., unsample) to produce a high-quality HR output $ I_{HR} $:
	\begin{equation}
		I_{HR} = Unsample(F_{SF}+F_{DF}).  
	\end{equation} 
        To achieve high efficiency, in the HR image reconstruction module, we use a convolution layer to refine features and a pixel-shuffle layer to upscale the final features to the HR image.

	\subsection{UFONE: A Deployment-Friendly Transformer Unit}
        To mitigate the adverse impact of reshaping operations during the inference stage, we propose a deployment-friendly transformer unit, namely UFONE.
 	Different from the networks designed for high-level vision tasks~(e.g., image classification and object detection), which obtain most semantic information from the global feature, SISR networks also need to exploit the local feature to reconstruct HR image with fine details and textures. \cite{yoo2020rethinking,ning2021uncertainty}. 
        Based on this observation, the proposed UFONE unfolds features once in the beginning, which greatly reduces the computational burden and memory usage caused by frequent feature reshaping. 
        Specifically, each UFONE consists of successive $ M $ Inner-patch Transformer Layers (ITL) and $ N $ Spatial-Aware Layers (SAL). 
 	The Inner-patch Transformer Layer (ITL) is aimed to extract potential structural information from the interior of each patch. 
        Following the ITL, a Spatial-Aware Layer (SAL) is proposed to capture long-range dependencies that are ignored in ITL. 
        Finally, a convolution layer is introduced to enhance the modeling capabilities of the UFONE. 
	
	\subsubsection{Inner-patch Transformer Layer (ITL)}
	The architecture of ITL is shown in Figure 4(a), which includes an Inner-patch Self-Attenton (ISA). 
	Specifically, for an input patch $ X_{L} \in \mathbb{R}^{C\times P\times P} $, the proposed ISA first reshapes it to the size of $ C\times P^{2} $, where $ P $ is the size of the local patch. 
	Then the corresponding \textit{Query} (Q), \textit{Key} (K), and \textit{Value} (V) are generated through a linear layer. 
        Following Restormer \cite{zamir2022restormer}, we use the L2 Normalization (L2-Norm) to normalize Query and Key. 
	The result corresponding to self-attention is reshaped to the size of $ C\times P\times P $ again and the output feature patch $ \hat{Y}_{L} \in \mathbb{R}^{C\times P\times P} $ can be obtained after a convolution layer. 
	Here, ISA can be formulated as: 
	\begin{gather}
		Y_{L} = ConV(SelfAtten(Q,K,V)_{r}), \nonumber \\
		SelfAtten(Q,K,V) = V\cdot Softmax(K\cdot Q/\alpha ), 
	\end{gather} 	
	where $ \alpha $ is a trainable scaling parameter to control the magnitude of the element-wise product of $ K $ and $ Q $ before applying the softmax function. 
	\vspace{-2mm}

	\subsubsection{Spatial-Aware Layer (SAL)} 
	Besides the Inner-patch Transformer Layer (ITL), the Spatial-Aware Layer (SAL) is another key design in our UFONE. 
	As shown in Figure \ref{details}(b), the proposed SAL adopts an efficient MetaFormer-style structure \cite{yu2022metaformer} that evolved from the transformer architecture. 
        To capture long-range dependencies within a large receptive field, we introduce a Spatial Dilated Attention (SDA) with Dilated Convolution (DConV) \cite{wang2018smoothed} to the SAL. 
	Specifically, we first feed the global input $ X $ into a convolution layer and split it into two parts ($ (X_{1} $ and $ X_{2}) $) in the channel dimension by a SimpleGate \cite{chen2022simple}. 
	Then, three successive dilated convolution layers are used to capture long-range dependencies from $ X_{1} $ and multiply with $ X_{2} $. 
	Finally, the features are passed to convolutional layers to produce output features. 
	Overall, the SAL process is defined as: 
	\begin{gather}
		Y = ConV(Atten(X_{1}, X_{2})), \nonumber \\
		Atten(X_{1}, X_{2}) = DConVs(X_{1})\cdot X_{2}. 
	\end{gather}

\begin{figure}[t]
        \setlength{\abovecaptionskip}{0pt}
        \setlength{\belowcaptionskip}{-1pt}
		\centering
		\includegraphics[width=\linewidth]{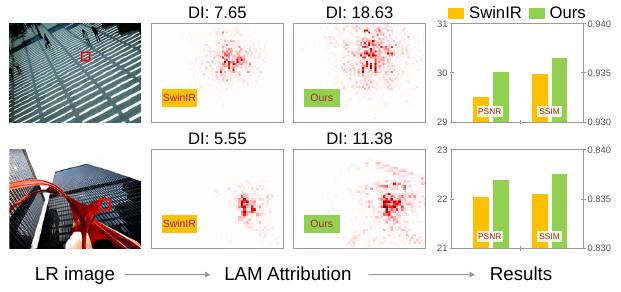}
		\caption{Comparison of the LAM attribution\cite{gu2021interpreting} and corresponding quantitative results for different SISR networks. LAM highlights the range of pixels involved in the reconstructing of SR results, quantified by the diffusion index (DI). A larger DI indicates that more pixels are involved.}
		\label{lam}
	\end{figure}

        To evaluate the effectiveness of the proposed SAL, we adopt LAM attribution ~\cite{gu2021interpreting} to determine the ability of capturing long-range dependencies of our method.
	As shown in Figure \ref{lam}, due to the large receptive field in the SAL, our DITN can capture long-range spatial information better than SwinIR \cite{liang2021swinir}. 
	As a result, our model can reconstruct HR images with higher PSNR/SSIM values. 
        \vspace{-3mm}

        \subsection{Deployment Optimization in Realistic Scenarios}
        \label{optimization}
        \vspace{-1mm}
        \subsubsection{Revisit LayerNorm layer.} 
        In addition to designing a deployment-friendly UFONE to eliminate redundant reshaping operations in the transformer, we further explored alternatives of LayerNorm \cite{ba2016layer}. 
        As mentioned in previous work \cite{li2022efficientformer,zhang2019root}, LayerNorm needs to dynamically handle the re-centering and re-scaling of the input and weight matrices, resulting in high latency and more memory usage on the deployment platform. 

	Given an input vertor of size $ H $: \textbf{\textit{X}} = $ (x_{1}, x_{1}, ..., x_{H}) $. 
	The output of a LayerNorm layer \textbf{\textit{Y}} can be written as: 
	\begin{equation}
		\textbf{\textit{Y}} = g \odot N(\textbf{\textit{X}})+b, ~~~~~
		N(\textbf{\textit{X}}) = \frac{\textbf{\textit{X}}-\mu}{\sigma} \in [-1,1], 
		\label{layernorm}
	\end{equation} 	
	where $ \odot $ denotes the element-wise product, 
	$ \mu $ and $ \sigma $ are the mean and standard statistics of \textbf{\textit{X}}, and
	gain $ g $ and bias $ b $ are parameters used to enhance expressive power of models. 
	To develop an efficient substitution for LayerNorm, a simple Tanh activation function is first used to imitate the normalization N in the Equation \ref{layernorm} and constrain the feature values to the range -1 to 1. 
	Then, a 1 $\times$ 1 convolution is followed to act as affine transformation on normalized vectors, just as $ g $ and $ b $ do in Equation \ref{layernorm}. 
	Therefore, a substitution layer for LayerNorm can be designed as: 
	\begin{equation}
		\textbf{\textit{Y}} = \textbf{\textit{W}} \odot \sigma (\textbf{\textit{X}})+\textbf{\textit{B}}, ~~~~~
		\sigma (\textbf{\textit{X}}) = Tanh(\textbf{\textit{X}}) \in [-1,1], 
	\end{equation} 	
	where \textbf{\textit{W}} and \textbf{\textit{B}} are the weight and bias of convolution. 
        \vspace{-2mm}

        \subsubsection{Operator Optimization in Self-Attention Layer.}
\begin{figure}[t]
        \setlength{\abovecaptionskip}{0pt}
        \setlength{\belowcaptionskip}{-1pt}
		\centering
		\includegraphics[width=\linewidth]{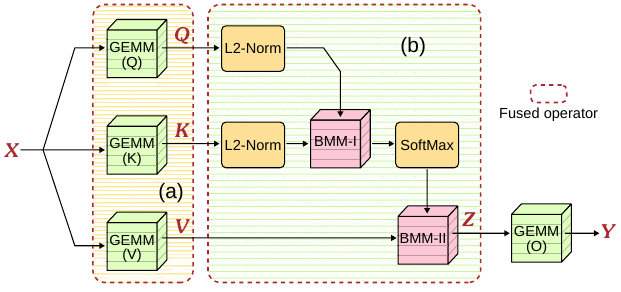}
		\caption{Details of the proposed operator fusion optimization strategy. We fuse the box-selected operators in the figure into a new operator.}
		\label{fused_op}
	\end{figure}

        Self-attention is the key layer of the transformer, which accounts for most computation in the transformer. 
        To improve the computational and deployment efficiency of the self-attention layer, we propose an effective operator fusion optimization strategy on the TensorRT-based deployment platform. 
        Since a single TensorRT instance can only serially execute a single General Matrix Multiplications (GEMM) operator, the original self-attention layer requires three serial GEMM operators to generate Q, K, and V respectively, as shown in Figure \ref{fused_op} (a), which not only fails to fully utilize computing resources but also seriously affects the throughput of data flow. 
        To this end, we fuse three GEMMs with the same channel configuration into a more computationally efficient triple-channel GEMM. 
        The outputs are written to GPU High Bandwidth Memory (HBM) corresponding to Q, K, and V respectively according to the offset. 

        Furthermore, five intermediate variables are activated in the following computation, as shown in Figure \ref{fused_op} (b), resulting in frequent HBM accesses. 
        To alleviate this problem, we introduce the FlashAttention\cite{dao2022flashattention}, which is an IO-aware exact attention algorithm. 
        Specifically, combined with the L2-Norm in self-attention, we develop an efficient FlashAttention in CUDA to achieve fine-grained control over memory access and merge all the attention operations into one GPU kernel. 
        Note that all CUDA kernels are registered to TensorRT through the plugins mechanism. 
        As a result, the proposed models run faster and with less memory access than standard self-attention.

	\section{Experiments}
 
	\subsection{Experimental Settings}

        \subsubsection{Network Configuration.} 
        We use three (i.e., $ K $=3) UFONEs to extract deep features from shallow features and the number of $ M $ and $ N $ is set to 4 by default. 
        The channel number $ C $ of the deep features is set to 60. 
        We adopt a single-head self-attention in ISA and apply self-attention across channels. 
	Since the Gated-Dconv Feed-forward Network (GDFN) \cite{zamir2022restormer} has shown great potential in controlling the feature transformation, we still use GDFN as the feed-forward network in the proposed ITL and SAL. 
	Based on the above configuration, we construct the DITN model. 
	We further obtain a lighter and more efficient model called DITN-Tiny by reducing the number of $ K $ from 3 to 1. 
	Furthermore, for deployment in realistic scenarios, we provide a Tensor-RT model named DITN-Real by substituting the NormLayer in DITN-Tiny with the proposed alternative in Section \ref{optimization} and adopting the operator fusion optimization. 
	In this paper, our experiments are conducted based on the above three models. 
 
        \subsubsection{Training Details.} 
	Similar to the most SISR methods \cite{liang2021swinir,zhang2021pffn,lu2022transformer}, we train DITN on 800 training images from DIV2K \cite{agustsson2017ntire}, and use the five benchmark datasets, including Set5 \cite{bevilacqua2012low}, Set14 \cite{zeyde2012single}, BSD100 \cite{martin2001database}, Urban100 \cite{huang2015single}, and Manga109 \cite{matsui2017sketch}, for quantitative and qualitative evaluation. 
	Besides, we adopt Adam optimizer \cite{kingma2014adam} ($ \beta_{1} $ = 0.9 and $ \beta_{2} $ = 0.99) and $ L1 $ loss to optimize our DITN. 
	All the images are randomly cropped to a size of 64 $\times$ 64. 
    We set the patch size in all Inner-patch Transformer Layers (ITL) to 8 $\times$ 8. 
	The initial learning rate is set to 2e-4. The batch size is 64. 
        \vspace{-2mm}
 
	\begin{table*}[t]
        \setlength{\abovecaptionskip}{0pt}
        \setlength{\belowcaptionskip}{-1pt}
		\caption{Quantitative results of state-of-the-art light-weighted SISR methods. Noted that all the efficiency proxies (i.e., Params, FLOPs, and Latency) are measured under the setting of upscaling HR images to 1280 $ \times $ 720 on all scales, and evaluated on Pytorch framework using a single Quadro RTX 8000 GPU. Best and second best results are \textcolor{red!60!black}{\textbf{highlighted}} and \underline{underlined}, respectively.}	
		\centering
        \renewcommand\arraystretch{0.8}
		\resizebox{0.99\textwidth}{!}{
			\begin{tabular}{c|c|c|cc|cc|cc|cc|cc||cc}
				\toprule
			    \multirow{2}{*}{Methods} & \multirow{2}{*}{Scale} & Latency(ms) & \multicolumn{2}{c|}{Set5 \cite{bevilacqua2012low}} & \multicolumn{2}{c|}{Set14 \cite{zeyde2012single}} & \multicolumn{2}{c|}{BSD100 \cite{martin2001database}} & \multicolumn{2}{c|}{Urban100 \cite{huang2015single}} & \multicolumn{2}{c||}{Manga109 \cite{matsui2017sketch}} & \#Param & \#FLOPs       \\ \cline{3-13} 
				& & Pytorch & PSNR & SSIM & PSNR & SSIM & PSNR & SSIM & PSNR & SSIM & PSNR & SSIM & (M) & (G)     \\ 
				\hline
				Bicubic & & - & 33.66 & 0.9299 & 30.24 & 0.8688 & 29.56 & 0.8431 & 26.88 & 0.8403 & 30.80 & 0.9339 & - & - 			\\
				CARN \cite{ahn2018fast} & & 57.0 & 37.76 & 0.9590 & 33.52 & 0.9166 & 32.09 & 0.8978 & 31.92 & 0.9256 & 38.36 & 0.9765 & 1.592 & 222.8 			\\
				LAINet \cite{xiao2021self} & & - & 37.94 & 0.9604 & 33.52 & 0.9174 & 32.12 & 0.8991 & 31.67 & 0.9242 & - & - & 0.237 & - 			\\
				IMDN \cite{hui2019lightweight} & & 44.9 & 38.00 & 0.9605 & 33.63 & 0.9177 & 32.19 & 0.8996 & 32.17 & 0.9283 & 38.88 & 0.9774 & 0.694 & 158.8 			\\
				LAPAR-A \cite{li2020lapar} & & 60.9 & 38.01 & 0.9605 & 33.62 & 0.9183 & 32.19 & 0.8999 & 32.10 & 0.9283 & 38.67 & 0.9772 & 0.548 & 171.0 			\\
				LatticeNet \cite{luo2020latticenet} & $\times$ 2 & 53.7 & 38.06 & 0.9607 & 33.70 & 0.9187 & 32.20 & 0.8999 & 32.25 & 0.9288 & - & - & 0.756 & 169.5 			\\
				PFFN \cite{zhang2021pffn} & & - & 38.07 & 0.9607 & 33.69 & 0.9192 & 32.21 & 0.8997 & 32.33 & 0.9298 & 38.89 & 0.9775 & 0.569 & 138.3 			\\ 
				ESRT \cite{lu2022transformer} & & - & 38.03 & 0.9600 & 33.75 & 0.9184 & 32.25 & 0.9001 & 32.58 & 0.9318 & \underline{39.12} & 0.9774 & 0.677 & - 						\\
				ELAN-Light \cite{zhang2022efficient} & & 278.6 & \textcolor{red!60!black}{\textbf{38.17}} & 0.9611 & \textcolor{red!60!black}{\textbf{33.94}} & \textcolor{red!60!black}{\textbf{0.9207}} & 32.30 & \underline{0.9012} & \underline{32.76} & \underline{0.9340} & 39.11 & \underline{0.9782} & 0.621 & 168.4 						\\
				SwinIR-Light \cite{liang2021swinir} & & 929.2 & 38.14 & \textcolor{red!60!black}{\textbf{0.9611}} & \underline{33.86} & \underline{0.9206} & \underline{32.31} & \underline{0.9012} & \underline{32.76} & \underline{0.9340} & \underline{39.12} & \textcolor{red!60!black}{\textbf{0.9783}} & 0.910 & 195.6 						\\
				DITN (Ours) & & 251.2 & \textcolor{red!60!black}{\textbf{38.17}} & \textcolor{red!60!black}{\textbf{0.9611}} & 33.79 & 0.9199 & \textcolor{red!60!black}{\textbf{32.32}} & \textcolor{red!60!black}{\textbf{0.9014}} & \textcolor{red!60!black}{\textbf{32.78}} & \textcolor{red!60!black}{\textbf{0.9343}} & \textcolor{red!60!black}{\textbf{39.21}} & 0.9781 & 1.120 & 228.3 		\\	
				\hline
				Bicubic &  & - & 30.39 & 0.8682 & 27.55 & 0.7742 & 27.21 & 0.7385 & 24.46 & 0.7349 & 26.95 & 0.8556 & - & -            \\
				CARN \cite{ahn2018fast} & & 31.5 & 34.29 & 0.9255 & 30.29 & 0.8407 & 29.06 & 0.8034 & 28.06 & 0.8493 & 33.50 & 0.9440 & 1.592 & 118.8 			\\
				LAINet \cite{xiao2021self} & & - & 34.26 & 0.9261 & 30.24 & 0.8404 & 29.04 & 0.8039 & 27.83 & 0.8453 & - & - & 0.237 & - 			\\
				IMDN \cite{hui2019lightweight} &  & 21.6 & 34.36 & 0.9270 & 30.32 & 0.8417 & 29.09 & 0.8046 & 28.17 & 0.8519 & 33.61 & 0.9445 & 0.703 & 71.5            \\
				LAPAR-A \cite{li2020lapar} &  & 42.8 & 34.36 & 0.9267 & 30.34 & 0.8421 & 29.11 & 0.8054 & 28.15 & 0.8523 & 33.51 & 0.9441 & 0.544 & 114.0             \\
				LatticeNet \cite{luo2020latticenet} & $\times$ 3 & 25.8 & 34.40 & 0.9272 & 30.32 & 0.8416 & 29.10 & 0.8049 & 28.19 & 0.8513 & - & - & 0.765 & 76.3           \\
				PFFN \cite{zhang2021pffn} & & - & 34.54 & 0.9282 & 30.42 & 0.8435 & 29.17 & 0.8062 & 28.37 & 0.8566 & 33.63 & 0.9455 & 0.558 & 69.1 			\\ 
				ESRT \cite{lu2022transformer} &  & - & 34.42 & 0.9268 & 30.43 & 0.8433 & 29.15 & 0.8063 & 28.46 & 0.8574 & 33.95 & 0.9455 & 0.770 & -         \\
				ELAN-Light \cite{zhang2022efficient} &  & 121.1 & 34.61 & 0.9288 & \textcolor{red!60!black}{\textbf{30.55}} & \underline{0.8463} & \underline{29.21} & 0.8081 & \textcolor{red!60!black}{\textbf{28.69}} & \underline{0.8624} & \underline{34.00} & \underline{0.9478} & 0.629 & 75.7            \\
				SwinIR-Light \cite{liang2021swinir} & & 313.2 & \underline{34.62} & \underline{0.9289} & 30.54 & \underline{0.8463} & 29.20 & \underline{0.8082} & 28.66 & \underline{0.8624} & 33.98 & \underline{0.9478}  & 0.918 & 87.2             \\
				DITN (Ours) & & 115.4 & \textcolor{red!60!black}{\textbf{34.63}} & \textcolor{red!60!black}{\textbf{0.9290}} & \textcolor{red!60!black}{\textbf{30.55}} & \textcolor{red!60!black}{\textbf{0.8467}} & \textcolor{red!60!black}{\textbf{29.23}} & \textcolor{red!60!black}{\textbf{0.8088}} & \underline{28.68} & \textcolor{red!60!black}{\textbf{0.8630}} & \textcolor{red!60!black}{\textbf{34.15}} & \textcolor{red!60!black}{\textbf{0.9482}} & 1.128 & 102.0         \\
				\hline
				Bicubic & & - & 28.42 & 0.8104 & 26.00 & 0.7027 & 25.96 & 0.6675 & 23.14 & 0.6577 & 24.89 & 0.7866  & - & -            \\
				CARN \cite{ahn2018fast} & & 25.1 & 32.13 & 0.8937 & 28.60 & 0.7806 & 27.58 & 0.7349 & 26.07 & 0.7837 & 30.47 & 0.9084 & 1.592 & 90.9 			\\
				LAINet \cite{xiao2021self} & & - & 32.12 & 0.8942 & 28.59 & 0.7810 & 27.55 & 0.7351 & 25.92 & 0.7805 & - & - & 0.263 & - 			\\
				IMDN \cite{hui2019lightweight} & & 13.4 & 32.21 & 0.8948 & 28.58 & 0.7811 & 27.56 & 0.7353 & 26.04 & 0.7838 & 30.45 & 0.9075 & 0.715 & 40.9          \\
				LAPAR-A \cite{li2020lapar} &  & 37.9 & 32.15 & 0.8944 & 28.61 & 0.7818 & 27.61 & 0.7366 & 26.14 & 0.7871 & 30.42 & 0.9074 & 0.659 & 94.0         \\
				LatticeNet \cite{luo2020latticenet} & $\times$ 4 & 16.9 & 32.18 & 0.8943 & 28.61 & 0.7812 & 27.57 & 0.7355 & 26.14 & 0.7844 & - & - & 0.777 & 43.6         \\
				PFFN \cite{zhang2021pffn} & & - & 32.36 & 0.8967 & 28.68 & 0.7827 & 27.63 & 0.7370 & 26.26 & 0.7904 & 30.50 & 0.9100 & 0.569 & 45.1 			\\ 
				ESRT \cite{lu2022transformer} &  & - & 32.19 & 0.8947 & 28.69 & 0.7833 & \underline{27.69} & 0.7379 & 26.39 & 0.7962 & 30.75 & 0.9100 & 0.751 & -          \\
				ELAN-Light \cite{zhang2022efficient} &  & 68.5 & 32.43 & 0.8975 & \underline{28.78} & \underline{0.7858} & \underline{27.69} & \underline{0.7406} & \textcolor{red!60!black}{\textbf{26.54}} & \underline{0.7982} & \underline{30.92} & 0.9150 & 0.640 & 43.2            \\
				SwinIR-Light \cite{liang2021swinir} &  & 176.7 & \underline{32.44} & \underline{0.8976} & 28.77 & \underline{0.7858} & \underline{27.69} &\underline{ 0.7406} & 26.47 & 0.7980 & \underline{30.92} & \underline{0.9151} & 0.929 & 49.6            \\
				DITN (Ours) &  & 64.5 & \textcolor{red!60!black}{\textbf{32.54}} & \textcolor{red!60!black}{\textbf{0.8988}} & \textcolor{red!60!black}{\textbf{28.86}} & \textcolor{red!60!black}{\textbf{0.7874}} & \textcolor{red!60!black}{\textbf{27.72}} & \textcolor{red!60!black}{\textbf{0.7420}} & \textcolor{red!60!black}{\textbf{26.54}} & \textcolor{red!60!black}{\textbf{0.8001}} & \textcolor{red!60!black}{\textbf{30.98}} & \textcolor{red!60!black}{\textbf{0.9153}}  & 1.139 & 58.1         \\
				
				\bottomrule
			\end{tabular}
		}
		
		\label{tab1}
	\end{table*}
 	\begin{figure*}[t]
		\centering
		\includegraphics[width=\linewidth]{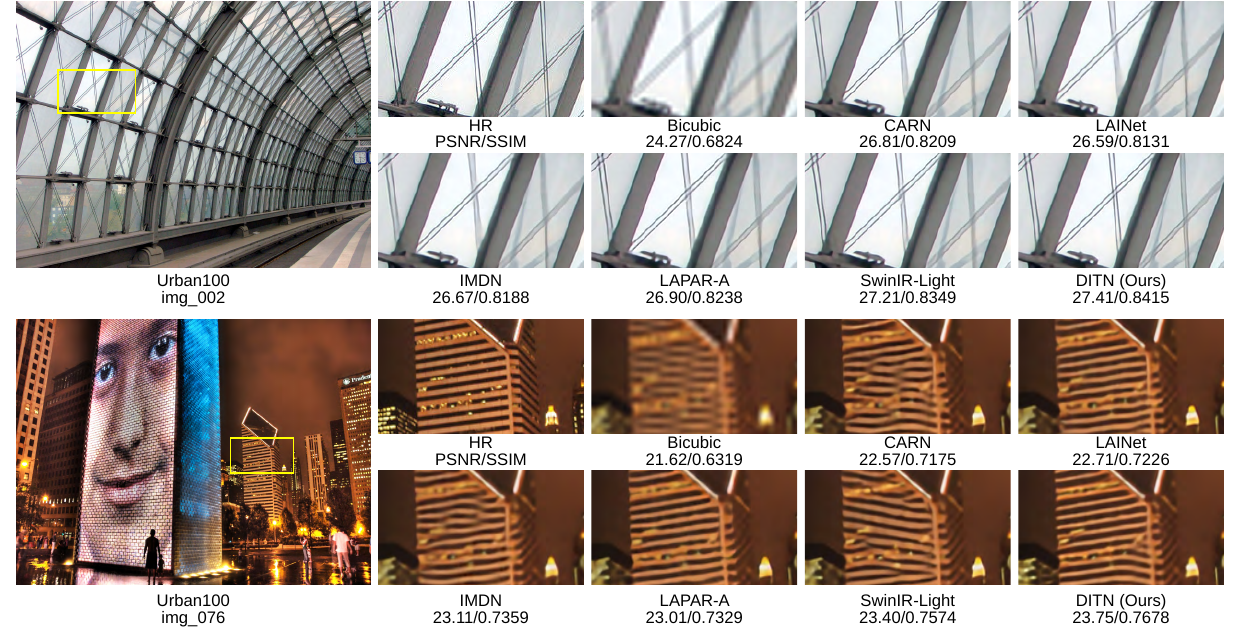}
		\caption{Visual comparison of state-of-the-art light-weighted SISR methods on Urban100 \cite{huang2015single} ($\times$ 4).}
		\label{images}
        \vspace{-1em}
	\end{figure*}

	\begin{table*}[t]
        \setlength{\abovecaptionskip}{0pt}
        \setlength{\belowcaptionskip}{-1pt}
		\caption{Quantitative results of different light-weighted SISR methods on five datasets. All the efficiency proxies follow the same setup as Table \ref{tab1}. Best and second best results are \textcolor{red!60!black}{\textbf{highlighted}} and \underline{underlined}, respectively.}
		\centering
        \renewcommand\arraystretch{0.85}
		\resizebox{0.99\textwidth}{!}{
			\begin{tabular}{c|c|c|cc|cc|cc|cc|cc|cc}
				\toprule
				\multirow{2}{*}{Methods} & \multirow{2}{*}{Scale} & \multicolumn{1}{c|}{Latency(ms)} & \multicolumn{2}{c|}{Set5 \cite{bevilacqua2012low}} & \multicolumn{2}{c|}{Set14 \cite{zeyde2012single}} & \multicolumn{2}{c|}{BSD100 \cite{martin2001database}} & \multicolumn{2}{c|}{Urban100 \cite{huang2015single}} & \multicolumn{2}{c|}{Manga109 \cite{matsui2017sketch}} & \#Param & \#FLOPs        \\ \cline{3-13} 
				&& Pytorch & PSNR & SSIM & PSNR & SSIM & PSNR & SSIM & PSNR & SSIM & PSNR & SSIM & (M) & (G)      \\ 
				\hline
				ELAN-Tiny \cite{zhang2022efficient} & & 95 & 37.95 & \textcolor{red!60!black}{\textbf{0.9605}} & 33.59 & 0.9182 & \underline{32.15} & \underline{0.8994} & 31.89 & 0.9266 & 38.40 & \textcolor{red!60!black}{\textbf{0.9769}} & 0.212 & 56.4 						\\
				SwinIR-Tiny \cite{liang2021swinir} & $\times$ 2 & 278 & 37.93 & \underline{0.9604} & 33.57 & 0.9179 & \underline{32.15} & 0.8992 & 31.95 & 0.9265 & 38.39 & \underline{0.9767} & 0.323 & 63.3 						\\
				DITN-Tiny (Ours) & & 85 & \underline{38.00} & \underline{0.9604} & \textcolor{red!60!black}{\textbf{33.70}} & \textcolor{red!60!black}{\textbf{0.9192}} & \textcolor{red!60!black}{\textbf{32.16}} & \textcolor{red!60!black}{\textbf{0.8995}} & \textcolor{red!60!black}{\textbf{32.08}} & \textcolor{red!60!black}{\textbf{0.9282}} & \textcolor{red!60!black}{\textbf{38.59}} & \textcolor{red!60!black}{\textbf{0.9769}} & 0.367 & 74.6		\\	
				DITN-Real (Ours) & & 78 & \textcolor{red!60!black}{\textbf{38.01}} & \textcolor{red!60!black}{\textbf{0.9605}} & \underline{33.65} & \underline{0.9184} & \textcolor{red!60!black}{\textbf{32.16}} & \textcolor{red!60!black}{\textbf{0.8995}} & \underline{31.96} & \underline{0.9273} & \underline{38.49} & \underline{0.9767} & 0.425 & 86.9 		\\	    
				\hline
				ELAN-Tiny \cite{zhang2022efficient} & & 43 & 34.23 & 0.9262 & 30.30 & 0.8416 & 29.04 & 0.8044 & 27.93 & 0.8484 & 33.19 & 0.9427 & 0.221 & 25.4 		\\	
				SwinIR-Tiny \cite{liang2021swinir} & $\times$ 3 & 95 & \underline{34.35} & \underline{0.9266} & \underline{30.35} & \underline{0.8428} & 29.07 & \underline{0.8051} & \underline{28.10} & \underline{0.8513} & 33.39 & \underline{0.9442} & 0.331 & 28.2 		\\	
				DITN-Tiny (Ours) & & 40 & \textcolor{red!60!black}{\textbf{34.38}} & \textcolor{red!60!black}{\textbf{0.9271}} & \textcolor{red!60!black}{\textbf{30.37}} & \textcolor{red!60!black}{\textbf{0.8435}} & \textcolor{red!60!black}{\textbf{29.10}} & \textcolor{red!60!black}{\textbf{0.8057}} & \textcolor{red!60!black}{\textbf{28.14}} & \textcolor{red!60!black}{\textbf{0.8529}} & \textcolor{red!60!black}{\textbf{33.56}} & \textcolor{red!60!black}{\textbf{0.9447}} & 0.376 & 33.2 		\\	
				DITN-Real (Ours) & & 36 & 34.33 & \underline{0.9266} & 30.33 & 0.8424 & \underline{29.08} & \underline{0.8051} & 28.06 & 0.8512 & \underline{33.44} & 0.9441 & 0.433 & 38.7 		\\	
				\hline
				ELAN-Tiny \cite{zhang2022efficient} & & 25 & 32.01 & 0.8921 & 28.49 & 0.7788 & 27.50 & 0.7342 & 25.86 & 0.7782 & 30.01 & \underline{0.9027}	& 0.231 & 14.5 	\\	
				SwinIR-Tiny \cite{liang2021swinir} & $\times$ 4 & 48 & 32.09 & 0.8933 & \underline{28.59} & \underline{0.7813} & 27.55 & 0.7356 & \underline{25.99} & 0.7819 & \underline{30.32} & \underline{0.9069} & 0.343 & 16.1 		\\	
				DITN-Tiny (Ours) & & 24 & \textcolor{red!60!black}{\textbf{32.16}} & \textcolor{red!60!black}{\textbf{0.8943}} & \textcolor{red!60!black}{\textbf{28.62}} & \textcolor{red!60!black}{\textbf{0.7825}} & \textcolor{red!60!black}{\textbf{27.59}} & \textcolor{red!60!black}{\textbf{0.7370}} & \textcolor{red!60!black}{\textbf{26.04}} & \textcolor{red!60!black}{\textbf{0.7853}} & \textcolor{red!60!black}{\textbf{30.40}} & \textcolor{red!60!black}{\textbf{0.9076}} & 0.387 & 18.9 		\\	
				DITN-Real (Ours) & & 22 & \underline{32.10} & \underline{0.8940} & \underline{28.59} & \underline{0.7813} & \underline{27.57} & \underline{0.7363} & \underline{25.99} & \underline{0.7837} & 30.29 & 0.9068 & 0.444 & 22.1 		\\					
				\bottomrule
			\end{tabular}
		}
		
		\label{tab2}
        \vspace{-1em}
	\end{table*}
	\begin{figure}[t]
        \setlength{\abovecaptionskip}{0pt}
        \setlength{\belowcaptionskip}{-1pt}
		\centering
		\includegraphics[width=0.7\linewidth]{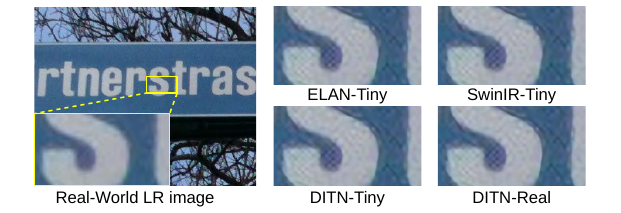}
		\caption{Visual comparison of different models on a real-world LR image from \cite{lugmayr2020ntire}. }
		\label{real}
        \vspace{-4.5mm}
	\end{figure}


	\subsection{Evaluation on the Training Platform}
	\label{experiments}
        
        To demonstrate the effectiveness of the architecture of DITN, we conduct a series of quantitative and qualitative experiments with several state-of-the-art SISR methods on the Pytorch framework. 
        \vspace{-2mm}
        
        \subsubsection{Quantitative Comparison.} 
        As shown in Table \ref{tab1}, we first evaluate our DITN against the following state-of-the-art light-weighted SISR methods: 
        Bicubic, CARN \cite{ahn2018fast}, LAINet \cite{xiao2021self}, IMDN \cite{hui2019lightweight}, LAPAR-A \cite{li2020lapar}, LatticNet \cite{luo2020latticenet}, PFFN \cite{zhang2021pffn}, ESRT \cite{lu2022transformer}, ELAN-Light \cite{zhang2022efficient}, and SwinIR-Light \cite{liang2021swinir}. 
        Note that real-world scenarios have strict requirements for the model used, e.g., < 2M parameters and < 200ms latency on the $ \times $ 4 scale. 
        Based on this consideration, we choose only the corresponding light version of ELAN \cite{zhang2022efficient} and SwinIR \cite{liang2021swinir} for further comparison. 
        The PSNR and SSIM computed in the YCbCr color space are used to quantify the difference.

        It can be found that the transformer-based methods occupy the best or second best places due to their self-attention mechanisms and content-adaptive flexibility. 
        Taking advantage of our efficient transformer unit UFONE, the proposed DITN achieves competitive results on almost all the five datasets and on all the three scaling factors (e.g., PSNR: \textbf{+0.15dB}, SSIM: \textbf{+0.0004} for $ \times $ 3 Manga109 and PSNR: \textbf{+0.10dB}, SSIM: \textbf{+0.0012} for $ \times $ 4 Set5). 
        In addition, the number of network parameters and FLOPs are also given as a reference, although they are not considered to correlate well with the complexity and efficiency of the network \cite{zhang2022efficient,xia2022trt}. 
        Compared with other transformer-based models, the proposed DITN has a slight increase in these two metrics, but still can achieve lower latency (\textbf{5\%$\sim $10\%} faster than ELAN-Tiny and \textbf{63\%$\sim $73\%} faster than SwinIR-Tiny) since it does not require frequent feature reshaping. 

        To evaluate the performance of the proposed method on extremely small size, we further report the quantitative results of the proposed DITN-Tiny and DITN-Real. 
        For fair comparison, we also obtain Swin-Tiny and ELAN-Tiny by reducing the corresponding model parameters to 1/3 times and retraining them with original codes and datasets. 
        As shown in Table \ref{tab2}, the proposed DITN-Tiny achieves the best PSNR/SSIM values in all comparisons (except the second best in $ \times $ 2 Set5). 
        While replacing LayerNorm with tanh activation function and 1$\times$1 convolution slightly degrades the performance than DITN-Tiny, the DITN-Real significantly speeds up the model's inference (\textbf{8\%$\sim $10\%} faster), and maintains consistent performance on real-world LR images (which will be analyzed in qualitative comparisons). 
        To further evaluate the efficiency of the proposed models on real-world LR images, we retrain four models (i.e.,Swin-Tiny, ELAN-Tiny, DITN-Tiny, and DITN-Real) with adversarial learning \cite{goodfellow2020generative} and our models can generate more accurate and natural image with fine details. 
        More experimental analysis are included in the supplementary.

	\begin{figure*}[t]
        \setlength{\abovecaptionskip}{-0pt}
        \setlength{\belowcaptionskip}{-1pt}
		\centering
		\includegraphics[width=\linewidth]{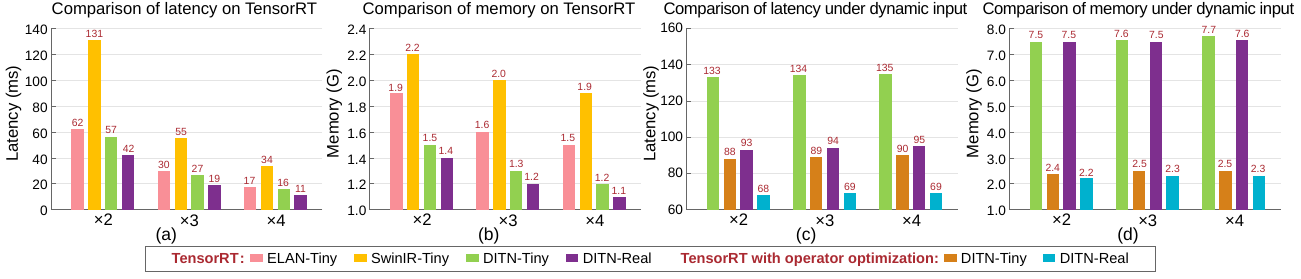}
		\caption{Quantitative comparison of deployment data of different transformer-based models on TensorRT. Note that all the deployment proxies in (a) and (b) are measured under the setting of upscaling HR images to 1280 $ \times $ 720, all the deployment proxies in (c) and (d) are measured under the setting of LR images to 960 $ \times $ 540, on a single NVIDIA A10 GPU.}
		\label{tensorRT}
        \vspace{-1mm}
	\end{figure*}

        \subsubsection{Qualitative Comparisons.} 
        We conduct qualitative comparisons of several SR images from Urban100 \cite{huang2015single} in Figure \ref{images}. 
        Urban100 is a challenging dataset that often used for vision comparison in previous works \cite{zhang2022efficient,lim2017enhanced,guo2020dual,zhang2021pffn}, involving various styles of urban scenes. 
        Note that since ELAN \cite{zhang2022efficient} has not released the pre-trained model for ELAN-Light, we can only make qualitative comparisons with SwinIR-Light in the transformer camp. 
        As we can see, the bicubic method yields highly blurred results on the two images, lacking clear texture structures. Utilizing the powerful learning capabilities of CNN and Transformer, other comparative methods (i.e., CARN \cite{ahn2018fast}, LAINet \cite{xiao2021self}, IMDN \cite{hui2019lightweight}, LAPAR-A \cite{li2020lapar}, and SwinIR-Light \cite{liang2021swinir}) are able to achieve clear results to varying degrees. Nevertheless, they still fail to reconstruct sharp structural textures (e.g., the stripe textures on the window in img\_002) and are susceptible to obvious distortions (e.g., the building outlines in img\_076). 
        These issues result in inaccurate reconstruction of latent details from LR images. In contrast, the proposed DITN succeeds in producing better results with clearer and sharper edges, which can deceive the ground truth to some extent. 
        Furthermore, we further verify the effectiveness of the proposed DITN-Real in processing real-world LR images in Figure \ref{real}. 
        It can be seen that DITN-Real can recover high-quality HR images consistent with DITN-Tiny, and both are better than the results of ELAN-Tiny and SwinIR-Tiny. 
        \vspace{-2mm}

        \subsection{Evaluation on the Deployment Platform}
        \label{deployment}
        To verify the effectiveness of our models on the deployment platform, we comprehensively analyze the performance of the proposed models under both static and dynamic inputs based on TensorRT. 

        First, we quantitatively evaluate the latency and memory usage of four tiny models (i.e., ELAN-Tiny, SwinIR-Tiny, DITN-Tiny, and DITN-Real) under static input configurations on TensorRT. 
        The results are shown in Figure \ref{tensorRT} (a) and (b). 
        Both latency and memory usage are efficiency metrics that provide more comprehensive feedback about the compute capacity and bandwidth on the deployment platform than parameters and FLOPs. 
        It can be seen that our models (DITN-Tiny and DITN-Real) have lower TensorRT latency and less memory usage compared to ELAN-Tiny and SwinIR-Tiny. 
        In particular, without LayerNorm, DITN-real achieves \textbf{26\%$\sim$ 31\%} TensorRT latency speedup and \textbf{6\%$\sim$ 8\%} memory savings than DITN-Tiny. 
        
        Since the practical SISR models need to handle arbitrary resolution inputs, it is also necessary to evaluate these models with dynamic input on the deployment platform. 
        To this end, we re-deploy these models with dynamic input configurations and quantitatively evaluate the latency and memory usage under dynamic input on both TensorRT and TensorRT with operator optimization, respectively. 
        The results are presented in Figure \ref{tensorRT} (c) and (d). 
        Noting that the unique operators in ELAN-Tiny and SwinIR-Tiny make them unavailable in the dynamic input configurations, we compare only DITN-Tiny and DITN-Real. 
        Specifically, we choose the LR of 960 $ \times $ 540 as inputs, and generate 1080P (1920 $ \times $ 1080), 2K (2880 $ \times $ 1620), and 4K (3840 $ \times $ 2160) outputs using $ \times $ 2, $ \times $ 3, and $ \times $ 4 models, respectively. 
        According to the charts, compared to DITN-Tiny, DITN-Real shows a huge lead with about \textbf{30\%} and \textbf{33.7\%} latency speedup on TensorRT and TensorRT with operator optimization respectively, making DITN-Real are more easier to deploy for real-world applications. 
        Furthermore, with the proposed operator fusion optimization strategy, (i): DITN-Tiny achieves about \textbf{33.5\%} latency speedup and about \textbf{68\%} memory savings; (ii): DITN-Real achieves about \textbf{26.8\%} latency speedup and about \textbf{70\%} memory savings. 

        Additionally, by referencing the proposed models on mobile devices, we find that our DITN-Real for real-world scenarios is 52\% faster than ELAN-Tiny and 42\% faster than DITN-Tiny (as shown in Table \ref{tab3}), and occupies only 36M of memory, which verifies the deployment feasibility of our model on edge/mobile devices. 

        \begin{table}[t]
        \setlength{\abovecaptionskip}{0pt}
        \setlength{\belowcaptionskip}{-2pt}
		\caption{Quantitative comparison of
  latency of different transformer-based models on CoreML. All these data are measured using an iPhone 14 under the setting of upscaling HR images to 1280 $ \times $ 720 and $ \times $ 4 scale. Note that used iPhone is equipped with an A16 processor and its NPU does not support the deployment of Swin-Tiny. Best result is \textcolor{red!60!black}{\textbf{highlighted}}.}
		\centering
        \tabcolsep=0.12cm
        \renewcommand\arraystretch{0.5}
		\begin{tabular}{c|cccc}
			\toprule
			CoreML & ELAN-Tiny & Swin-Tiny & DITN-Tiny & DITN-Real\\
			\hline
			Latency(ms) & 168 & - & 138 & \textcolor{red!60!black}{\textbf{80}} \\
			\bottomrule
		\end{tabular}
		\label{tab3}
        \vspace{-4mm}
	\end{table}

	\subsection{Ablation Study}
        To built an efficient and deployment-friendly transformer unit for light-weighted SISR task, we introduce the UFONE (i.e., UnFolding ONce is Enough) to our model. 
        In this section, we conduct ablation studies to better demonstrate the effectiveness of UFONE. 
        Specifically, we evaluate the impact of different combinations of ITL and SAL in each UFONE on SR results. 
        The results are presented in Table \ref{tab4}. 
        For the rationality of the experiment, we fixed the total number of ITL and SAL to be 8. 
        If we consider using only ITL (i.e., M=8, N=0) or SAL (i.e., M=0, N=8) in UFONE to extract the deep features of the image, we find that the corresponding PSNR values are low. 
        This is because that both local structural features and long-range dependencies are necessary for the SISR task. 
        In addition, we also conducted comparative experiments for different number of combinations of ITL and SAL. 
        The results show that our model can go to the best effect when M=4 and N=4. 
        More ablation studies on ISA and SDA can be found in the supplementary. 

        \begin{table}[t]
        \setlength{\abovecaptionskip}{0pt}
        \setlength{\belowcaptionskip}{-2pt}
		\caption{Ablation study of different number configurations of ITL and SAL. M denotes the number of ITL and N denotes the number of SAL. All experiments are evaluated on Manga109 \cite{matsui2017sketch} ($\times$2). Best result is \textcolor{red!60!black}{\textbf{highlighted}}.}
		\centering
        \renewcommand\arraystretch{0.5}
		\begin{tabular}{c|c}
			\toprule
			Configurations  & PSNR (dB)\\
			\hline
			M = 8, N = 0 & 38.39\\
			M = 0, N = 8 & 38.44\\
			M = 6, N = 2 & 38.58\\
			M = 2, N = 6 & 38.57\\
			\hline
			M = 4, N = 4 (Ours) & \textcolor{red!60!black}{\textbf{38.67}} \\
			\bottomrule
		\end{tabular}
		\label{tab4}
        \vspace{-4mm}
	\end{table}


	\section{Conclusion}
	Starting from a novel perspective of transferring all the transformer components to the patch level, we introduced a deployment-friendly transformer unit UFONE (i.e., UnFolding ONce is Enough) and presented a Deployment-friendly Inner-patch Transformer Network (DITN) for SISR. 
        By utilizing the Inner-patch Transformer Layer (ITL) and Spatial-Aware Layer (SAL) to extract features in patch view and spatial view respectively, DITN significantly and steadily improves the SISR performance. 
        In addition, we have further boost the efficiency of the proposed DITN on the deployment platform by optimizing specific layers and exploring an operator fusion strategy for self-attention layer. 
        Extensive experiments show that our models can achieve outstanding performance on various standard benchmark datasets and different platforms. 
        Furthermore, it is noteworthy that the proposed UFONE and deployment-oriented optimization strategies are theoretically applicable to similar transformer architectures and tasks as well. 

        \section{Acknowledgments}
        This work was supported in part by the National Key Research and Development Program of China under Grant 2022YFB3303800, in part by National Major Science and Technology Projects of China under Grant 2019ZX01008101. 

\bibliographystyle{ACM-Reference-Format}
\balance
\bibliography{sample-base}

\end{document}


\title{Supplementary Material for "Unfolding Once is Enough: A Deployment-Friendly Transformer Unit for Super-Resolution"}
\renewcommand{\shorttitle}{Unfolding Once is Enough: A Deployment-Friendly Transformer Unit for Super-Resolution}

\ccsdesc[500]{Computing methodologies~Reconstruction}
\ccsdesc[300]{Computing methodologies~Image processing}
\ccsdesc[100]{Computing methodologies~Computational photography}

\keywords{Image super-resolution, Efficient transformer, Lightweight network, Model deployment}

\maketitle

\section{Implementation Details}

From the perspective of transferring all the transformer components to the patch level, we present a Deployment-friendly Inner-patch Transformer Network (DITN) for SISR, which includes three models: DITN, DITN-Tiny, and DITN-Real. 
Here, we first present more training details of these models, as a supplement to Section 4.1.2 of the main paper. 
Specifically, all the $\times$ 2 models are trained for a total of 500K iterations. 
The initial learning rate is set to 2e-4 and halved at [250K, 400K, 450K, 475K]. 
For the $\times$ 3 and $\times$ 4 models, we initialize the models with corresponding $\times$ 2 weights and train for 250K iterations. 
The initial learning rate is set to 1e-4 and halved at [125K, 200K, 225K, 237.5K]. 
We optimize all models using L1 loss. 
Since the goal of SISR is the ultimate application in the real-world scenarios, we further retrain four models (e.g., Swin-Tiny, ELAN-Tiny, DTN-Tiny, and DTN-Real) with adversarial learning \cite{goodfellow2020generative}. 
For fair comparison, we obtain Swin-Tiny and ELAN-Tiny by reducing the corresponding models (Swin-Light \cite{liang2021swinir} and ELAN-Light \cite{zhang2022efficient}) to 1/3 times. 
Specifically, we train these models on 800 training images from DIV2K \cite{agustsson2017ntire} and 2650 training images from Flickr2K \cite{timofte2017ntire}. 
To obtain more realistic LR image synthesis, we use the same degradation model as BSRGAN \cite{zhang2021designing}, which consists of randomly shuffled blur, downsampling, noise degradations, etc. 
The models are trained for 500K iterations with a initial learning rate of 1e-4. 
The learning rate is halved at [250K, 400K, 450K, 475K]. 
We optimize these models with a weighted combination of L1 loss, VGG perceptual loss, and least square PatchGAN loss \cite{isola2017image} with weights \{1, 1, 0.1\}, respectively.

To evaluate the efficiency of the proposed models on real-world LR images, we use the DPEDiphone-crop \cite{lugmayr2020ntire} and RealSRSet \cite{zhang2021designing} as the test sets. 
Among them, DPEDiphone-crop is provided in the Real-World Super-Resolution Challenge of NTIRE 2020, 
and RealSRSet consists of 20 real-world images either downloaded from the internet or chosen directly from previous test sets \cite{ignatov2017dslr,zhang2018ffdnet,martin2001database,matsui2017sketch}. 
In all cases, we use $\times$ 4 upsampling. 
The specific experimental evaluation results are presented in Section \ref{real-world}. 

\section{More Ablation Studies}
In this section, we further conduct ablation studies to demonstrate the effectiveness of the proposed single-head Inner-patch Self-Attention (ISA) in the Inner-patch Transformer Layer (ITL) and Spatial Dilated Attention (SDA) in Spatial-Aware Layer (SAL). 

\subsection{Effectiveness of the ISA} 
To design a lightweight and deployment-friendly transformer layer to reconstruct the local structural information from LR images, the proposed ITL embeds a single-head Inner-patch Self-Attention (ISA). 
Here, we verify its effectiveness through a series of comparative experiments. 
Specifically, we set the head number of ISA to 1, 2, 4, and 6, respectively, to train the model, and the results are reported in Table \ref{tab-isa}. 
Compared with typical multi-head self-attention, our single-head ISA scheme is more effective for the SISR task.

\begin{table}[t]
\setlength{\abovecaptionskip}{-0pt}
\setlength{\belowcaptionskip}{-1pt}
\caption{Ablation study of different number of attention heads in Inner-patch Self-Attention (ISA). All experiments are evaluated on Manga109 \cite{matsui2017sketch} ($\times$2). Best result is \textcolor{red!60!black}{\textbf{highlighted}}.}
\centering
\begin{tabular}{c|cc}
    \toprule
    Head number & PSNR (dB) & Latency (ms) \\
    \hline
    6 & 38.59 & 253.6 \\
    4 & 38.62 & 252.5 \\
    2 & 38.63 & 254.3 \\
    \hline
    1 (Ours) & \textcolor{red!60!black}{\textbf{38.67}} & \textcolor{red!60!black}{\textbf{251.2}} \\
    \bottomrule
\end{tabular}
\label{tab-isa}
\vspace{-3mm}
\end{table}

\begin{table}[t]
\setlength{\abovecaptionskip}{-0pt}
\setlength{\belowcaptionskip}{-1pt}
\caption{Ablation study of different convolution kernel sizes and dilation rates in Spatial Dilated Attention (SDA). K denotes the size of convolution kernel and D denotes the size of dilation rate. All experiments are evaluated on Manga109 \cite{matsui2017sketch} ($\times$2). Best result is \textcolor{red!60!black}{\textbf{highlighted}}.}
\centering
\begin{tabular}{c|c}
    \toprule
    Configurations & PSNR (dB) \\
    \hline
    K = 5, D = 2 & 38.47 \\
    K = 9, D = 4 & 38.61 \\
    \hline
    K = 7, D = 3 (Ours) & \textcolor{red!60!black}{\textbf{38.67}} \\
    \bottomrule
\end{tabular}
\label{tab-sda}
\vspace{-3mm}
\end{table}

\subsection{Effectiveness of the SDA} 
In addition to the Inner-patch Transformer Layer (ITL), the proposed UFONE also includes a Spatial-Aware Layer (SAL) to capture potentially useful long-range dependencies that are neglected in the ITL. 
To obtain a large receptive field, we embed an efficient Spatial Dilated Attention (SDA) based on dilated convolutions \cite{wang2018smoothed}. 
To verify the influence of different convolution kernel sizes and dilation rates on the performance of the model in SDA, we perform a series of comparative experiments, and the results are reported in Table \ref{tab-sda}. 
It can be observed that small receptive field (i.e., K = 5, D = 2) is unfavorable for capturing long-range dependencies due to the high resolution of intermediate features. 
Moreover, when we use a large receptive field (i.e., K = 9, D = 4), the model tends to ignore important dependencies between the neighboring patches, leading to inaccurate results. 
Therefore, we choose the combination (i.e., K = 7, D = 3) with higher PSNR as the final parameter of the model.

\section{Evaluations on L1-based models}
\begin{figure*}[t]
    \centering
    \includegraphics[width=\linewidth]{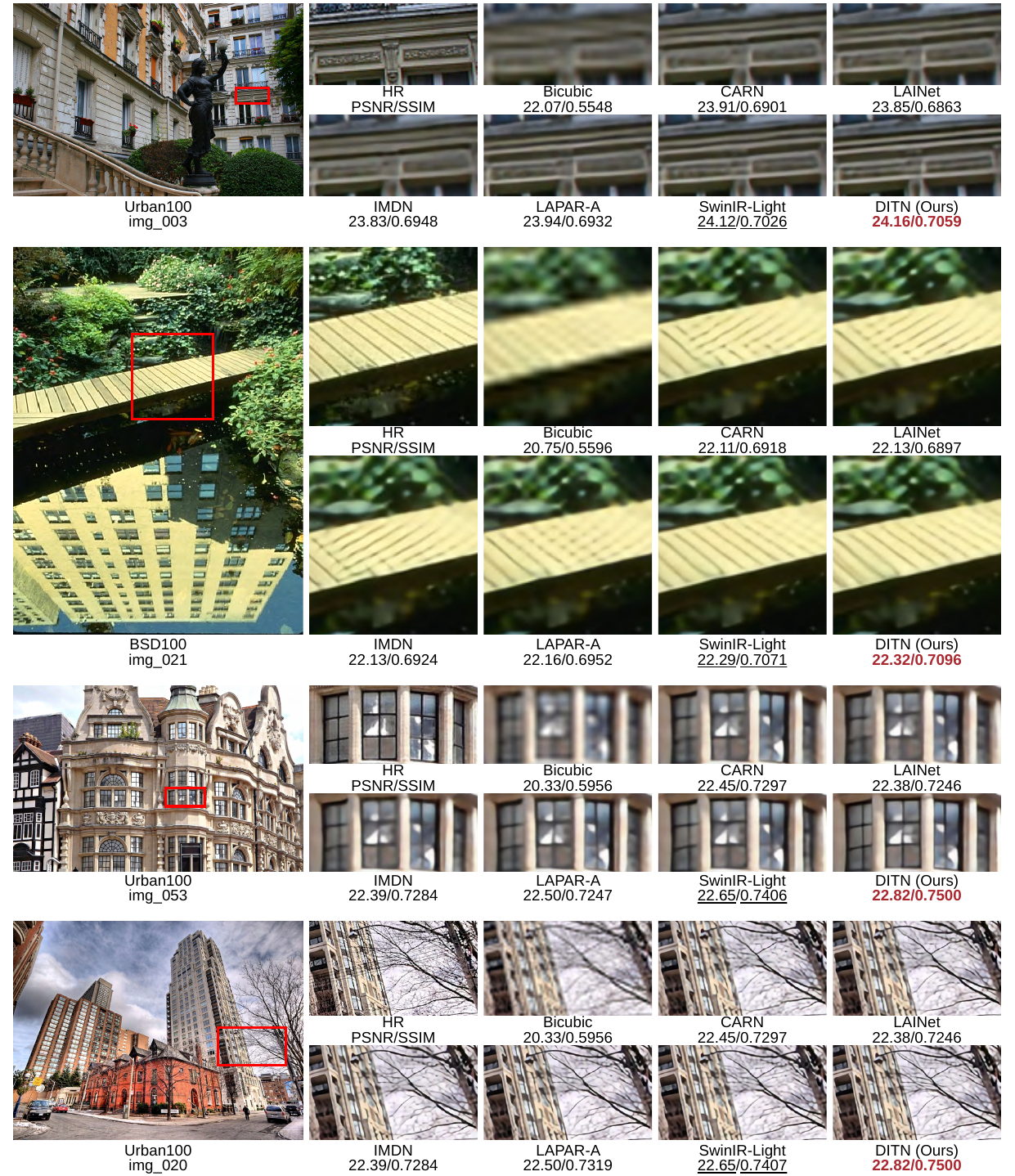}
    \caption{Visual comparison of the proposed DITN with other state-of-the-art light-weighted SISR methods on Urban100 \cite{huang2015single} and BSD100 \cite{martin2001database} ($\times $ 4). Best and second best PSNR/SSIM are \textcolor{red!60!black}{\textbf{highlighted}} and \underline{underlined}, respectively.}
    \label{supp-base}
\end{figure*}

\begin{figure*}[tt]
    \centering
    \includegraphics[width=\linewidth]{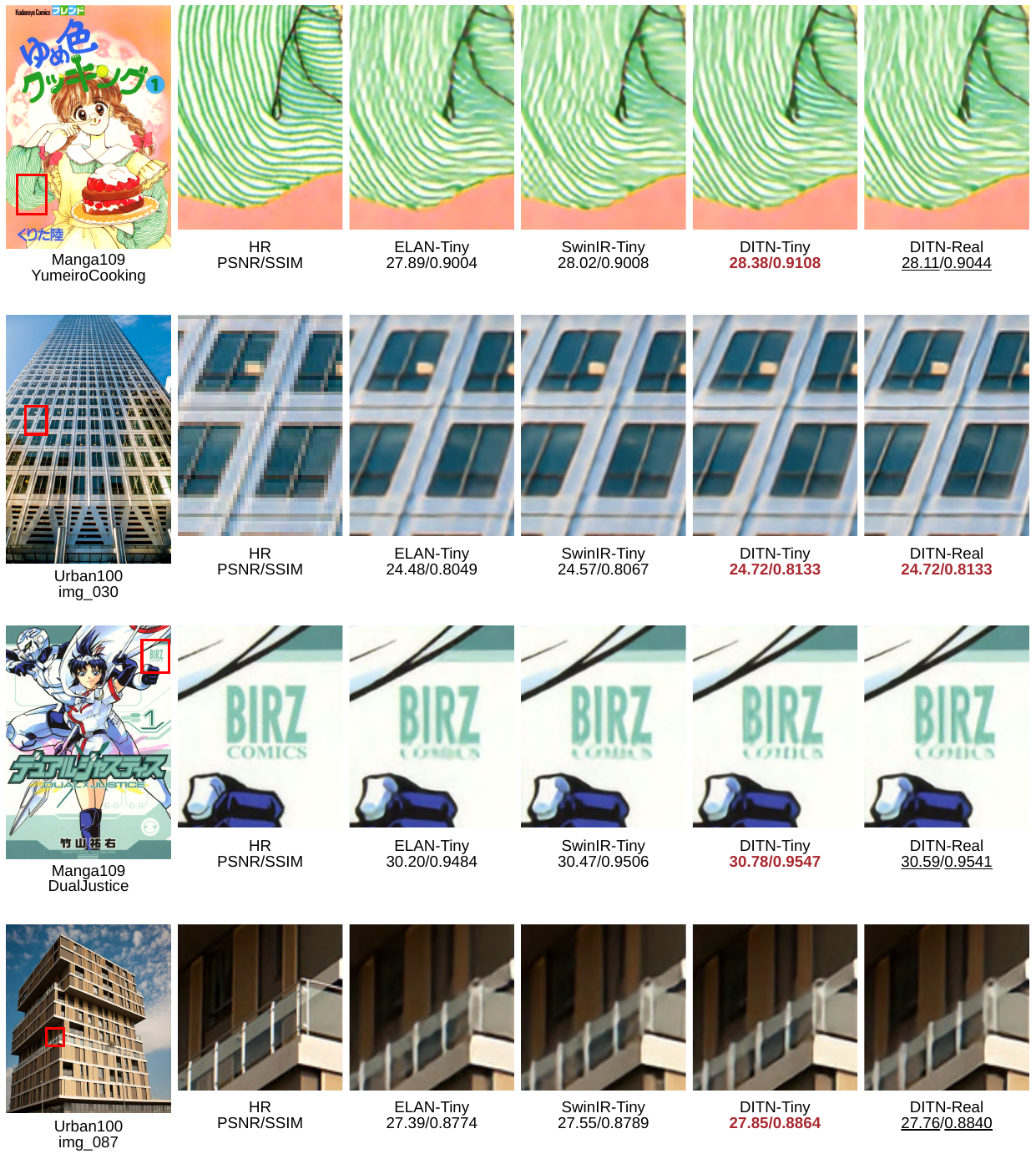}
    \caption{Visual comparison of the proposed DITN-Tiny and DITN-Real with other transformer-based tiny methods on Manga109 \cite{matsui2017sketch} and Urban100 \cite{huang2015single} ($\times $ 4). Best and second best PSNR/SSIM are \textcolor{red!60!black}{\textbf{highlighted}} and \underline{underlined}, respectively.}
    \label{supp-tiny}
\end{figure*}

In this section, we provide more visual evaluation results to complement the experimental analysis in the main paper. 
Specifically, as shown in Figure \ref{supp-base}, we present more visual evaluation results of the proposed DITN with state-of-the-art light-weighted SISR methods, corresponding to Table 1 and Figure 7 of the main paper. 
Additionally, as shown in Figure \ref{supp-tiny}, we also present more visual evaluation results of the proposed DITN-Tiny and DITN-Real with other transformer-based tiny models, corresponding to Table 2 of the main paper. 
Note that the corresponding PSNR and SSIM in all results are given from a quantitative point of view as a reference. 
All these results show the superiority of our models in handling diverse LR images. 

\section{Evaluations on GAN-based models}
\label{real-world}

\begin{table}[t]
\setlength{\abovecaptionskip}{-0pt}
\setlength{\belowcaptionskip}{-1pt}
\caption{The no-reference NIQE \cite{mittal2012making}, NRQM \cite{ma2017learning}, and PI \cite{blau20182018} results of different methods on the DPEDiphone-crop \cite{lugmayr2020ntire} and RealSRSet \cite{zhang2021designing} test sets. Best and second best results are \textcolor{red!60!black}{\textbf{highlighted}} and \underline{underlined}, respectively.}
\centering
\resizebox{0.48\textwidth}{!}{
\begin{tabular}{c|ccc|ccc}
    \toprule
    \multicolumn{1}{c|}{Methods} & \multicolumn{3}{c|}{DPEDiphone-crop \cite{lugmayr2020ntire}} & \multicolumn{3}{c}{RealSRSet \cite{zhang2021designing}} \\ \cline{2-7}
    \textbf{(+GAN)} & NIQE $\downarrow$ & NRQM $\uparrow$ & PI $\downarrow$ & NIQE $\downarrow$ & NRQM $\uparrow$ & PI $\downarrow$ \\
    \hline
    ELAN-Tiny  & 6.27 & 4.61 & 6.11 & \textcolor{red!60!black}{\textbf{5.90}} & 4.41 & 5.87 \\
    SwinIR-Tiny & \textcolor{red!60!black}{\textbf{5.33}} & \textcolor{red!60!black}{\textbf{5.26}} & \textcolor{red!60!black}{\textbf{5.18}} & \underline{5.92} & \textcolor{red!60!black}{\textbf{4.94}} & \textcolor{red!60!black}{\textbf{5.69}} \\
    DITN-Tiny & \underline{5.83} & 4.75 & \underline{5.74} & 5.95 & 4.72 & 5.75 \\
    DITN-Real & 6.45 & \underline{4.88} & 5.98 & 6.12 & \underline{4.93} & \underline{5.71} \\
    \bottomrule
\end{tabular}}
\label{tab-no}
\vspace{-3mm}
\end{table}

\begin{figure*}[t]
    \centering
    \includegraphics[width=\linewidth]{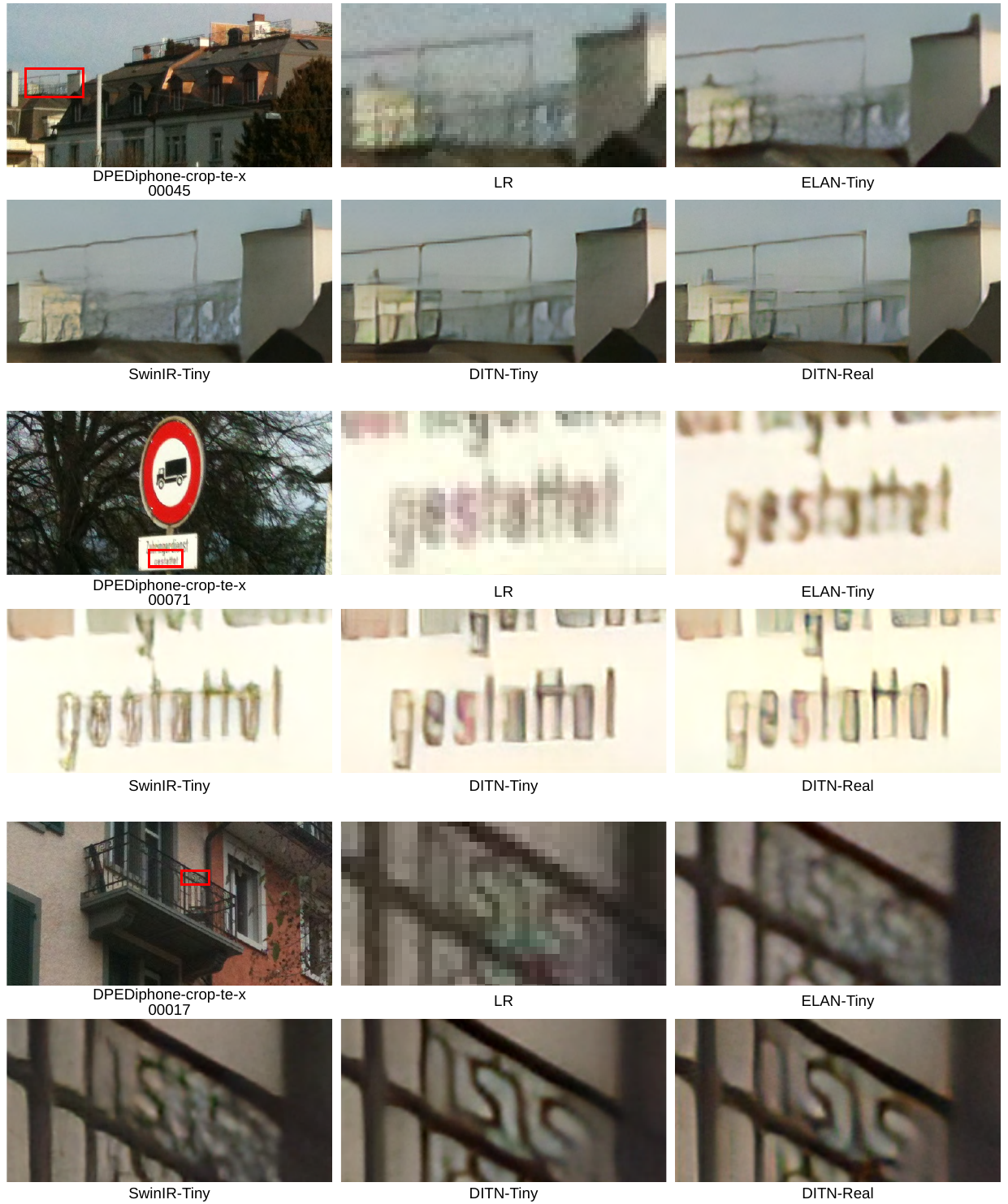}
    \caption{Visual comparison of the proposed DITN-Tiny (+GAN) and DITN-Real (+GAN) with other transformer-based tiny methods in real-world LR images of DPEDiphone-crop \cite{lugmayr2020ntire} ($\times $ 4).}
    \label{supp-iphone}
\end{figure*}

\begin{figure*}[t]
    \centering
    \includegraphics[width=\linewidth]{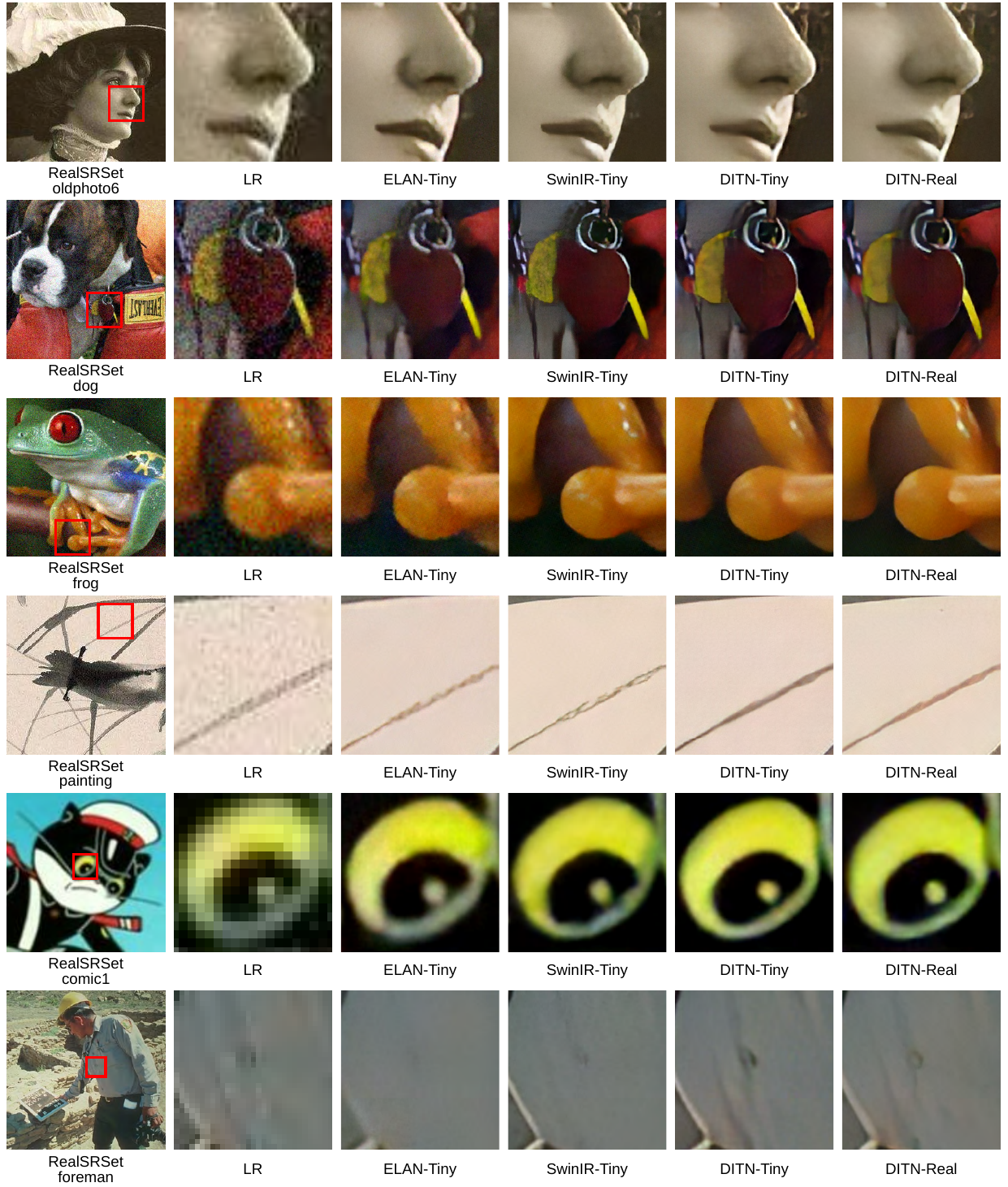}
    \caption{Visual comparison of the proposed DITN-Tiny (+GAN) and DITN-Real (+GAN) with other transformer-based tiny methods in real-world LR images of RealSRSet \cite{zhang2021designing} ($\times $ 4).}
    \label{supp-real}
\end{figure*}

In this section, we verify the effectiveness of the models trained with GAN on the real-world LR test set. 
Due to the lack of available ground truth HR images of the real-world LR images, we employ three non-reference image quality assessment metrics, i.e., NIQE \cite{mittal2012making}, NRQM \cite{ma2017learning}, and PI \cite{blau20182018}, for quantitative evaluation. 
The results are reported in Table \ref{tab-no}. 
Note that although SwinIR-Tiny achieves the promising quantitative values, their results fail to reconstruct accurate geometric details and result in unsatisfactory artifacts, which can be confirmed by the visual examples shown in Figure \ref{supp-iphone} and Figure \ref{supp-real}. 
This is because these existing non-reference metrics miss updates on the GAN-based data \cite{jinjin2020pipal} and thus do not match well with perceived visual quality in this setting \cite{lugmayr2020ntire,zhang2021designing,jinjin2020pipal}. 
To this end, we evaluate the performance of these GAN-based models based on the qualitative results, the examples of which are shown in Figure \ref{supp-iphone} and Figure \ref{supp-real}. 
It can be obviously seen that our models, i.e., DITN-Tiny (+GAN) and DITN-Real (+GAN), can produce better visual quality with clear and sharp edges. 
Specifically, thanks to the efficient combination of the successive Inner-patch Transformer layers (ITL) and Spatial-Aware Layers (SAL) in UFONE, the proposed models can accurately reconstruct both the text information on the sign (the second image in Figure \ref{supp-iphone}) and the outline of the railing (the third image in Figure \ref{supp-iphone}), while ensuring the consistency of the spatial content information. 
As shown in Figure \ref{supp-real}, our models also show greater advantages in reconstructing sharper facial contours, clearer joints, and more accurate painting textures, while other compared models (i.e., ELAN-Tiny (+GAN) and SwinIR-Tiny (+GAN)) may result in unsatisfactory artifacts. 
It is worth mentioning that by replacing LayerNorm in DITN-Tiny with fast and efficient substitution of the Tanh activation function and 1$\times$1 convolution, DITN-Real also outperforms compared models in both artifact removal and enhancement of image texture details.

\bibliographystyle{ACM-Reference-Format}
\bibliography{sample-base}